\documentclass[journal]{IEEEtran}
\usepackage{cite}
\usepackage{amsmath,amssymb,amsfonts}
\usepackage{algorithmic}
\usepackage{graphicx}
\usepackage{textcomp}
\usepackage{xcolor}
\usepackage{booktabs}
\usepackage{subcaption}
\usepackage{hyperref}
\usepackage{balance}

\def\BibTeX{{\rm B\kern-.05em{\sc i\kern-.025em b}\kern-.08em
    T\kern-.1667em\lower.7ex\hbox{E}\kern-.125emX}}
\usepackage{tikz,xcolor,hyperref}
\definecolor{lime}{HTML}{A6CE39}
\DeclareRobustCommand{\orcidicon}{%
	\begin{tikzpicture}
	\draw[lime, fill=lime] (0,0) 
	circle [radius=0.16] 
	node[white] {{\fontfamily{qag}\selectfont \tiny ID}};
	\draw[white, fill=white] (-0.0625,0.095) 
	circle [radius=0.007];
	\end{tikzpicture}
	\hspace{-2mm}
}
\foreach \x in {A, ..., Z}{%
	\expandafter\xdef\csname orcid\x\endcsname{\noexpand\href{https://orcid.org/\csname orcidauthor\x\endcsname}{\noexpand\orcidicon}}
}
\protect

\begin{document}
\title{Paradigm selection for Data Fusion \\ of SAR and Multispectral Sentinel data \\applied to Land-Cover Classification}
\author{\IEEEauthorblockN{Alessandro Sebastianelli\orcidA{}\IEEEauthorrefmark{1},
Maria Pia Del Rosso\orcidB{}\IEEEauthorrefmark{1},}
\thanks{\IEEEauthorrefmark{1} Departement of Engineering, University of Sannio, Benevento, Italy. A. Sebastianelli (email: sebastianelli@unisannio.it), M. P. Del Rosso (email: mpdelrosso@unisannio.it) and S. L. Ullo (email: ullo@unisannio.it).}
\IEEEauthorblockN{Pierre Philippe Mathieu\IEEEauthorrefmark{2},}
\thanks{\IEEEauthorrefmark{2} Head of $\Phi$-lab Explore Office, European Space Agency, Frascati, Italy. P. P. Mathieu (email: pierre.philippe.mathieu@esa.int)}
\IEEEauthorblockN{Silvia L. Ullo\orcidC{}\IEEEauthorrefmark{1}
}
}

\IEEEtitleabstractindextext{
\begin{abstract}
Data fusion is a well-known technique, becoming more and more popular in the  Artificial Intelligence for Earth Observation (AI4EO) domain mainly due to its ability of rein\nobreak forcing AI4EO applications by combining multiple data sources and thus bringing better results. 
On the other hand, like other methods for satellite data analysis, data fusion itself is also benefiting and evolving thanks to the integration of Artificial Intelligence (AI). In this letter, four data fusion paradigms, based on Convolutional Neural Networks (CNNs), are analyzed and implemented. The goals are to provide a systematic procedure for choosing the best data fusion framework, resulting in the best classification results, once the basic structure for the CNN has been defined, and to 
 help interested researchers in their work when data fusion applied to remote sensing is involved.  
The procedure has been validated for land-cover classification but it can be transferred to other cases.
\end{abstract}

}
\maketitle
\IEEEdisplaynontitleabstractindextext
\IEEEpeerreviewmaketitle
\vspace{-15pt}
\section{Introduction}
Data fusion techniques, integrated with Machine or Deep Learning algorithms, are applied to satellite data of different domains in several use cases, that have included in the last years Sentinel-1 SAR data and Sentinel-2 multispectral data from the Copernicus mission. Data fusion of SAR and optical data has gained great interest since collecting many surface spectral signatures as answers at the different frequencies can result into a wider knowledge with respect to the case in which a single domain is considered. In literature, several examples can be found, where fusion of Sentinel-1 and Sentinel-2 data is accomplished in different ways for specific applications. For instance, in \cite{he2018multi} the authors combined time-series of Sentinel-1 and Sentinel-2 data to simulate optical images, while specific Earth Observation (EO) data fusion applications related to urban area monitoring are presented in \cite{tavares2019integration} and \cite{benedetti2018sentinel}, and  related to bio-masses and vegetation  in  \cite{attarzadeh2018synergetic, nuthammachot2020combined, navarro2019integration, heckel2020predicting}. Conversely, authors of \cite{bioresita2019fusion} and \cite{kaplan2018sentinel} have applied data fusion to monitor and map water surfaces and wetlands. Land cover mapping has been also explored in \cite{steinhausen2018combining} and \cite{ienco2019}, where the focus has been on the monsoon-region in the first case, while the combined use of multi-temporal and multi-modal data fusion has been considered in the second one.

With respect to the state of the art, where the approach is more application-centric, the key concept of this letter is to explore the importance of the data-fusion framework selection, once the dataset and the structure of the AI-based architecture has been defined. Namely, this letter aims to provide a systematic procedure for choosing the best data fusion framework, which will bring better results in the specific case of application.


The effectiveness of the proposed approach has been ver\nobreak ified by specializing the procedure on the AI4EO land-cover classification, but it is transferable to other cases by its nature. 

\section{Data and Methods}
The method proposed in this letter involves the use of Machine Learning (ML) algorithms, more specifically Deep Learning (DL) models based on CNNs \cite{goodfellow2016deep, book, lecun2015deep, kim2017convolutional}, structured by following classical data fusion frameworks. 

In the following sections, both the dataset and the data fusion architectures developed upon a common CNN structure will be explained in details, also by showing the achieved results. It is worth to highlight that the dataset and the data fusion architectures have been both developed from scratch and made available on Git-Hub for further analysis and investigation \cite{code}.

\vspace{-10pt}
\subsection{Dataset}
The dataset is composed of Sentinel-1 SAR data (in dual polarization) and  Sentinel-2 multispectral data, retrieved for the same region and clustered in five land-cover classes: $1)$ city, $2)$ coastline, $3)$ lake, $4)$ river and $5)$ vegetation.

Let \textbf{X}$^1_{lat,lon}\in \mathbb{R}^{W\times H \times P}$ be a single Sentinel-1 SAR acquisition, with a width $W$, a height $H$, acquired in $P$ different types of polarization and let
\textbf{X}$^2_{lat,lon}\in \mathbb{R}^{W\times H \times B}$ be a single Sentinel-2 multispectral acquisition, with a width $W$, a height $H$ and a number of bands defined by $B$, both acquired in the same geographical region defined by latitude $lat$ and longitude $lon$ values. Through the tool proposed in  ~\cite{sebastianelli2020automatic}, a set of Sentinel-1 and Sentinel-2 acquisitions has been downloaded from the Google Earth Engine (GEE) catalog ~\cite{gorelick2017google}, which contains the full Sentinel-1 archive of Ground Ranged Detected (GRD) products and the Sentinel-2 archive of Level-2A multi-spectral products. 

As already underlined, the proposed procedure for data fusion paradigm selection has been applied to the land-cover classification but it can be transferred to other cases. Three data fusion frameworks  based on a common CNN structure have been analyzed. For one of these, two different versions have been later discussed, by bringing the total number of considered  frameworks to four.  They have been trained in a supervised way, and so both the input and the ground truth needed to  be available, where as  inputs a Sentinel-1 acquisition, a Sentinel-2 acquisition or the aggregation of these two has been  used, depending on the data fusion paradigm, while as ground truth the vector \textbf{y}$_{lat,lon}\in \mathbb{R}^{C}$ (with  $C$ denoting the number of classes),  defined using the equations (1a)-(1e), has been utilized:
\begin{subequations}
    \begin{align}
        city:\       class\ 1 &\longrightarrow [1.00\ 0.00\ 0.00\ 0.00\ 0.00]\\
        coastline:\  class\ 2 &\longrightarrow [0.00\ 1.00\ 0.00\ 0.00\ 0.00]\\
        lake:\       class\ 3 &\longrightarrow [0.00\ 0.00\ 1.00\ 0.00\ 0.00]\\
        river:\      class\ 4 &\longrightarrow [0.00\ 0.00\ 0.00\ 1.00\ 0.00]\\
        vegetation:\ class\ 5 &\longrightarrow [0.00\ 0.00\ 0.00\ 0.00\ 1.00]
    \end{align}
    \label{eqn:onehot}
\end{subequations}
Namely, through the above equations  the class labels (e.g. city, river, etc.) are traduced into a vector   by following the one-hot encoding notation. Each entry of the class label represents the probability that an image belongs to that class. Therefore, unlike  the ground truth that is built to be perfect (each label must have only one entry equals to $1.00$ and the remaining are equal to $0.00$), the model can  return an output whose vector can be equal for instance to $[0.85\ 0.05\ 0.02\ 0.03\ 0.10]$, meaning that the input is classified at $85\%$ as class 1, at $5\%$ as class 2, at $2.5\%$ as class 3, etc. 

The images downloading and the label assignment processes have been repeated an arbitrary number of times (this choice will define the size of the final dataset) by selecting the  latitude $lat$ and longitude $lon$ values distributed over the land surface of interest, and considering the land cover classes previously defined. Inputs and  ground truths are then grouped into vectors, in order to set up the dataset as expressed through equations (\ref{eqn:vectors}a) and (\ref{eqn:vectors}b):
\begin{subequations}
    \begin{align}
        \Vec{\mathbf{X}}_{inputs} &= \begin{cases}&[\mathbf{X}^1_{lat_0,lon_0},\dots,\mathbf{X}^1_{lat_N,lon_N}]\\ &[\mathbf{X}^2_{lat_0,lon_0},\dots,\mathbf{X}^2_{lat_N,lon_N}]\end{cases}\\
        \Vec{\mathbf{y}}_{ground\_ truths} &=[\mathbf{y}_{lat_0,lon_0},\dots,\mathbf{y}_{lat_N,lon_N}]
    \end{align}
    \label{eqn:vectors}
\end{subequations}
where $\Vec{\mathbf{X}}_{inputs} \in \mathbb{R}^{N\times W \times H \times (P,B)}$ and $\Vec{\mathbf{y}}_{ground\_ truths} \in \mathbb{R}^{N\times C}$ with $N$ representing the number of pair of Sentinel-1 and Sentinel-2 acquisitions on the Earth surface.

Finally, the dataset is divided into three sub-datasets,  the training dataset $\Vec{\mathbf{X}}_{inputs}^t, \Vec{\mathbf{y}}_{ground\_ truths}^t \in \mathbb{R}^{M\times W \times H \times (P,B)}$ and  $\mathbb{R}^{M\times C}$ respectively, the validation dataset $\Vec{\mathbf{X}}_{inputs}^v, \Vec{\mathbf{y}}_{ground\_ truths}^v \in \mathbb{R}^{Q\times W \times H \times (P,B)}$ and $\mathbb{R}^{Q\times C}$,  the testing dataset $\Vec{\mathbf{X}}_{inputs}^t, \Vec{\mathbf{y}}_{ground\_ truths}^t \in \mathbb{R}^{S\times W \times H \times (P,B)}$ and $\mathbb{R}^{S\times C}$, with $M=85\%$ of $N$ , $Q=10\%$ of $N$ and $S=5\%$.

The final dataset contains $500$ pairs of Sentinel-1 and Sentinel-2 acquisitions ($100$ samples for each class). Using the training-validation-testing split factor above specified, the resulting training dataset is composed of $425$ samples, the validation dataset of $50$ samples and  the testing dataset of $25$ samples. These datasets are subjected to data augmentation for artificially increasing their actual size, through transformations of the starting image (e.g. rotation, crops, etc.). The result is a four-times larger dataset:   $1700$ samples for training, $200$ for validation and $100$ for testing. A version of the proposed dataset has been made available  open access online \cite{code}.
\vspace{-8pt}
\subsection{Analyzed approaches}
\begin{figure}[!ht]
    \centering
    \includegraphics[scale=0.11]{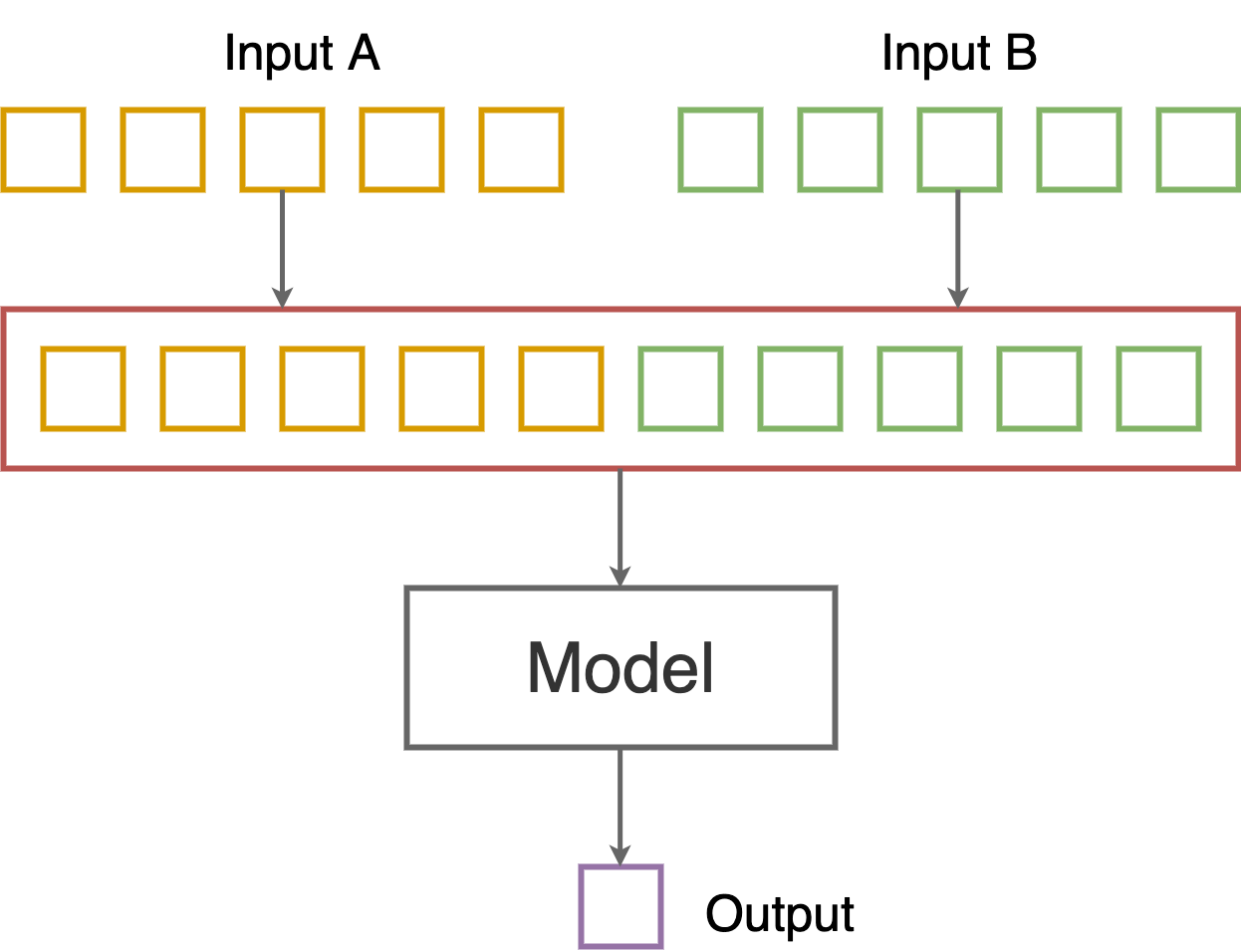}
    \caption{Early Data Fusion Paradigm}
    \label{fig:early}
\end{figure}

\begin{figure}[!ht]
    \centering
    \includegraphics[scale=0.11]{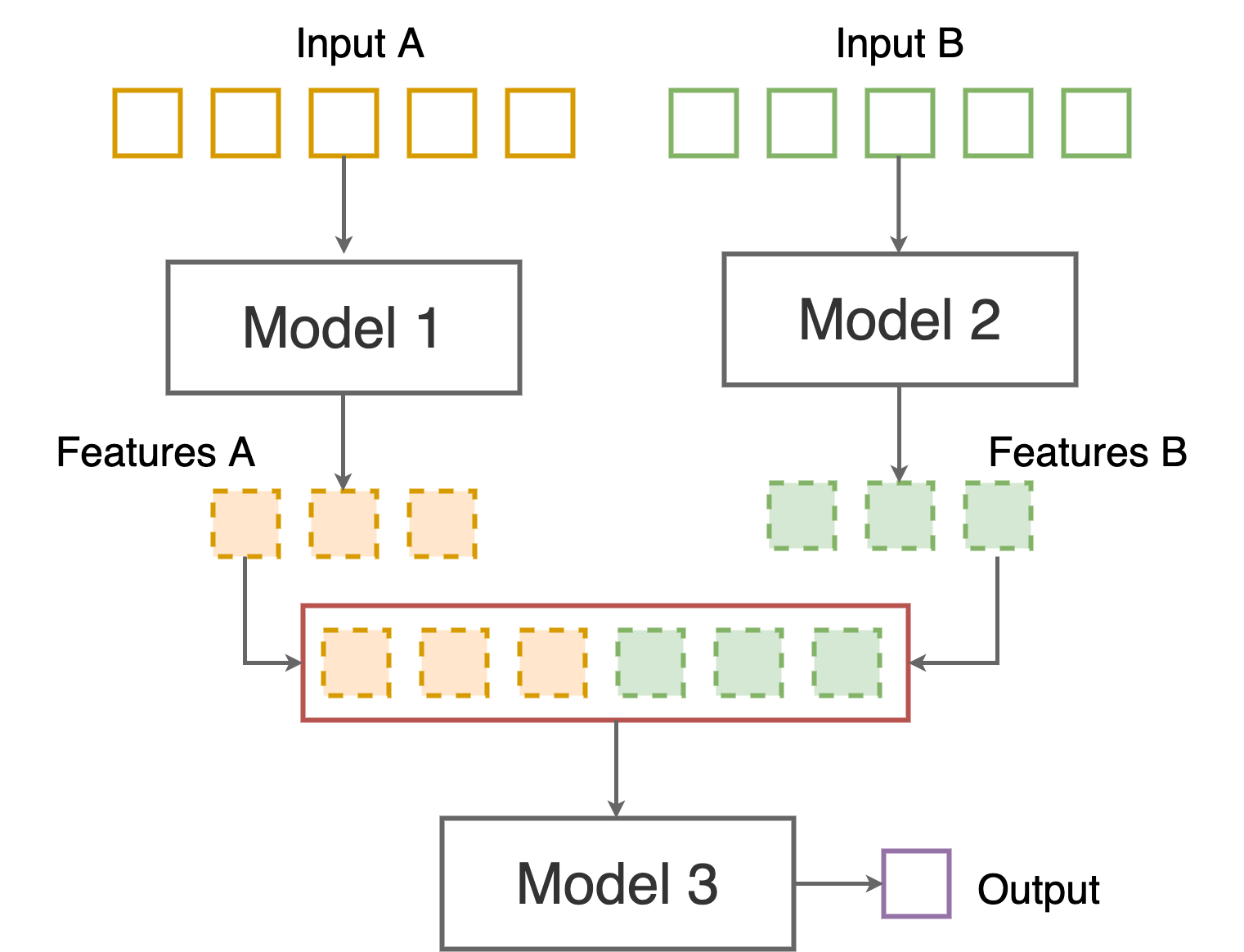}
    \caption{Joint Data Fusion Paradigm}
    \label{fig:joint}
\end{figure}

\begin{figure}[!ht]
    \centering
    \includegraphics[scale=0.11]{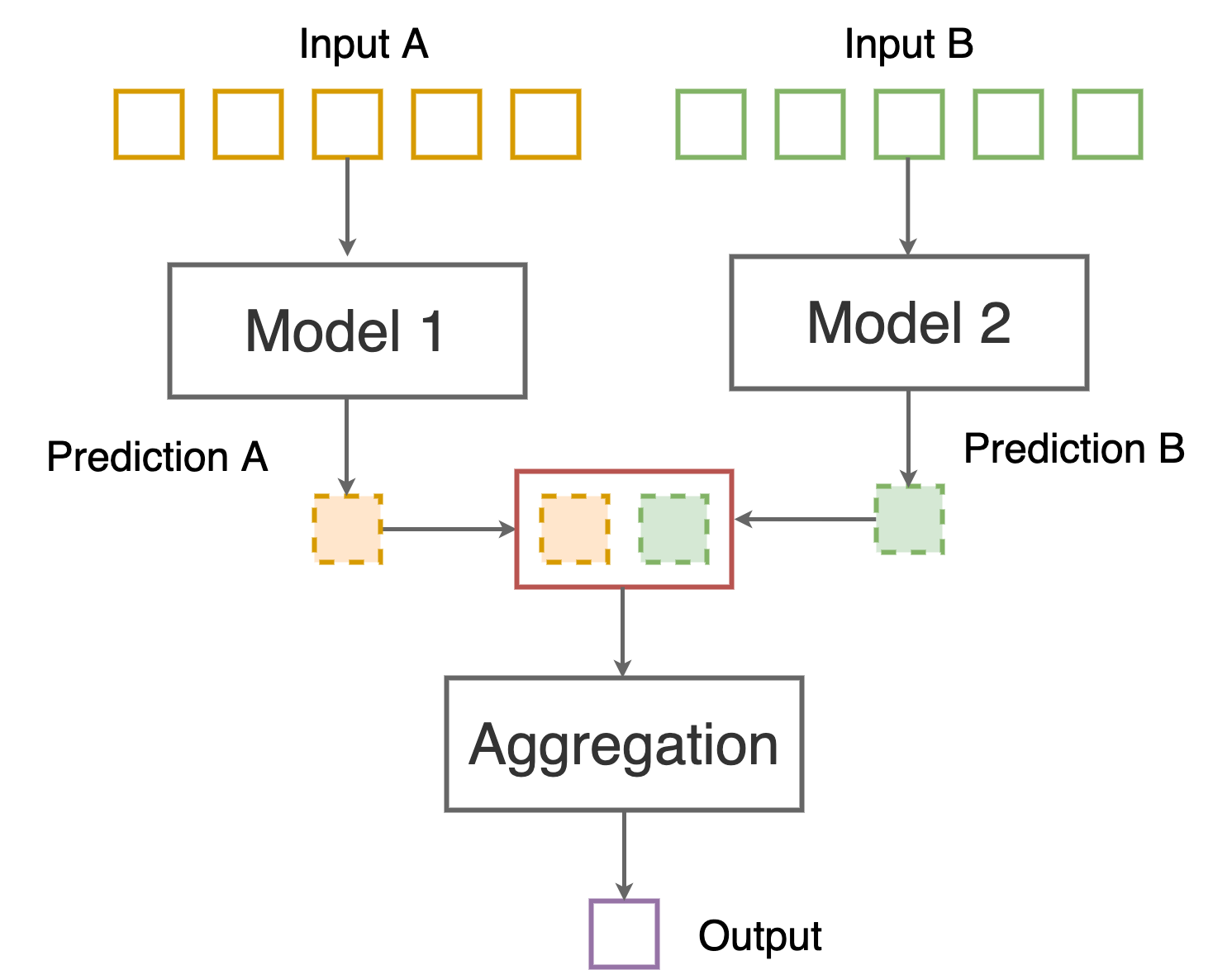}
    \caption{Late Data Fusion Paradigm}
    \label{fig:late}
\end{figure}

In this letter, three typical data fusion frameworks are analyzed: $1)$ Early Data Fusion, $2)$ Joint Data Fusion and $3)$ Late Data Fusion \cite{hall2001multisensor, liggins2017handbook, hall2004mathematical}. The Early Data Fusion paradigm consists in the use of one model, as shown in Fig. \ref{fig:early}, that takes the aggregation of \textit{Input A} and \textit{Input B} as input and returns a proper output.
Conversely, the Joint Data Fusion paradigm consists in the use of three models, as shown in Figure \ref{fig:joint}, where \textit{Model 1} and \textit{Model 2} are used for feature extraction, respectively from \textit{input A} and \textit{input B}, and the third model, \textit{Model 3},  combines these features to calculate the final output. 
On the contrary, the Late Data Fusion paradigm combines the outputs, \textit{Prediction A} and \textit{Prediction B}, of two models, \textit{Model 1} and \textit{Model 2}, to calculate the output, as shown in Fig. \ref{fig:late}.

\begin{figure*}[!ht]
    \centering
    \includegraphics[width=1.75\columnwidth]{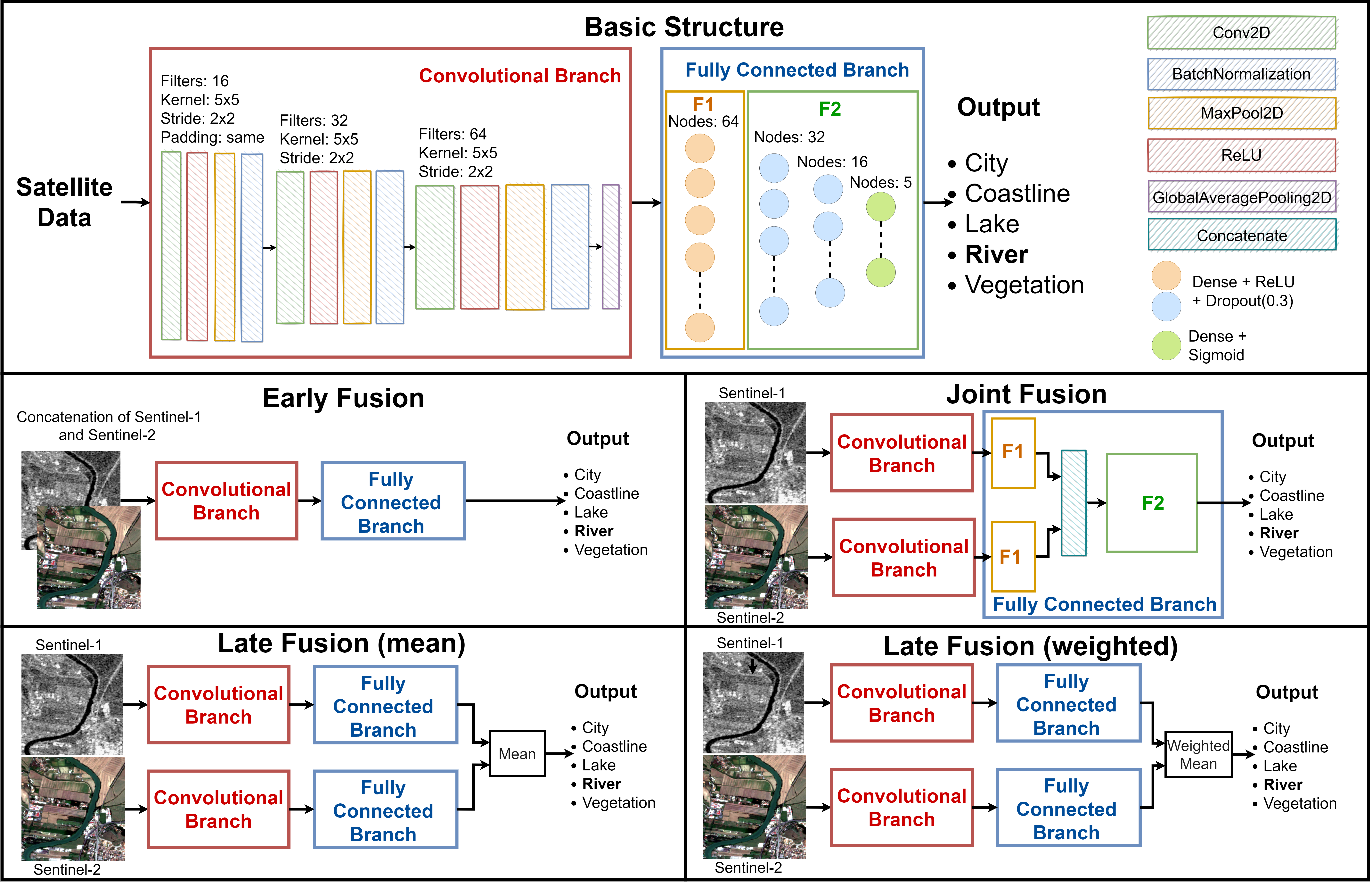}
    \caption{Proposed models architecture: Basic CNN structure (top row), Early Fusion (middle row - left), Joint Fusion (middle row - right), Late Fusion Mean Strategy (bottom row - left) and Late Fusion Weighted Strategy (bottom row - right). The figure shows as an example of classification for the "river" class.}
    \label{fig:model}
\end{figure*}

Regarding the Late Fusion paradigm,  two aggregation strategy were analyzed, one based on the simple mean and the other based on the weighted combination of prediction A and prediction B, by bringing the final number of the examined data fusion frameworks to four.

By starting from all the paradigms above explained  and combining them with the basic CNN structure presented in \ref{fig:model}, the four data fusion models have been designed. It is worth to highlight that the basic CNN structure is a classifier composed of two fundamentals branch: the Convolutional and the Fully Connected one. The Early Fusion paradigm takes as input the aggregation of a Sentinel-1 and Sentinel-2 image and its structure is the exact replica of the basic CNN structure. The fusion in this case  takes place both at beginning (image fusion) but also in the overall CNN (features combination). The Joint Data Fusion paradigm is composed of two separated convolutional branches, each  handling or the Sentinel-1 or the Sentinel-2 image, that later converge in one fully connected branch.  The fusion takes place at features' level. The two Late Fusion paradigms considered in this study differ only for the aggregation strategy (mean or weighted), while they are both composed of two separated basic CNNs and the fusion take places for both at the end. Moreover,  while the aggregation strategy is straightforward in the first case (mean), it is worth to specify that when the  weighted strategy is adopted, the aggregation is done in accordance with the   equation (3),  where $prediction_A$ and $prediction_B$ are the output of \textit{Model A} and \textit{Model B} respectively, expressed with the one-hot encoding notation defined before, while  $\alpha_x$ and $\beta_x$ are the weights used to combine the two predictions as shown in the Table \ref{tab:weights}. 
\begin{equation}
    \centering
    \begin{split}
    \mathbf{output} =
    &[\alpha_1, \alpha_2, \alpha_3, \alpha_4, \alpha_5] * prediction_A +\\
    &[\beta_1, \beta_2, \beta_3, \beta_4, \beta_5] * prediction_B
    \end{split}
\end{equation}

\begin{table}[!ht]
    \centering
    \resizebox{0.5\columnwidth}{!}{%
    \begin{tabular}{cccc}
    \toprule
	Parameter & Value & Parameter & Value\\
	\midrule
	$\alpha_1$ & 0 & $\beta_1$ & 1 \\
	$\alpha_2$ & 1 & $\beta_2$ & 0 \\
	$\alpha_3$ & 1 & $\beta_3$ & 0 \\
	$\alpha_4$ & 1 & $\beta_4$ & 0 \\
	$\alpha_5$ & 0 & $\beta_5$ & 1 \\
    \bottomrule
    \end{tabular}
    }
    \caption{Late Data Fusion: weights values}
    \label{tab:weights}
\end{table}
The selection of the weights highly depends on the performances of the two separated CNNs that compose the Late Fusion model.
\vspace{-13pt}
\section{Results and Discussion}
\begin{table*}[!ht]
    \centering
    \resizebox{1.85\columnwidth}{!}{%
    \begin{tabular}{c|ccccc|c|ccccc}
    \toprule
    \textbf{(A)} & & & \textbf{Confusion Matrix} & & & \textbf{(B)} & & & \textbf{Confusion Matrix} &  \\
    & & &    \textbf{Sentinel-2}    & & & & & &   \textbf{Sentinel-1}    &  \\
    \toprule
    & City & Coastline & Lake & River & Vegetation & & City & Coastline & Lake & River & Vegetation \\
    \midrule
    City       & 0.9  & 0.0  & 0.02 & 0.0   & 0.081 & City       & 0.88 & 0.0  & 0.0   & 0.12 & 0.0  \\
    Coastline  & 0.0  & 0.85 & 0.15 & 0.0   & 0.0   & Coastline  & 0.0  & 0.9  & 0.09  & 0.01 & 0.0  \\
    Lake       & 0.0  & 0.21 & 0.64 & 0.1   & 0.051 & Lake       & 0.0  & 0.15 & 0.71  & 0.14 & 0.0  \\
    River      & 0.0  & 0.0  & 0.32 & 0.66  & 0.02  & River      & 0.0  & 0.02 & 0.059 & 0.92 & 0.0  \\
    Vegetation & 0.03 & 0.0  & 0.02 & 0.099 & 0.85  & Vegetation & 0.3  & 0.0  & 0.0   & 0.16 & 0.54 \\
    \toprule
    \textbf{(C)} & & & \textbf{Confusion Matrix} & & & \textbf{(D)} & & & \textbf{Confusion Matrix} &  \\
    & & &    \textbf{Early Fusion}    & & & & & &   \textbf{Joint Fusion}    &  \\
    \toprule
    & City & Coastline & Lake & River & Vegetation & & City & Coastline & Lake & River & Vegetation \\
    \midrule
    City       & 0.89 & 0.0    & 0.0  & 0.0  & 0.11  & City       & 0.98 & 0.0   & 0.0  & 0.0  & 0.02   \\
    Coastline  & 0.0  & 0.78   & 0.17 & 0.02 & 0.03  & Coastline  & 0.0  & 0.81  & 0.13 & 0.06 & 0.0    \\
    Lake       & 0.0  & 0.11   & 0.65 & 0.21 & 0.03  & Lake       & 0.0  & 0.071 & 0.78 & 0.14 & 0.01   \\
    River      & 0.0  & 0.0099 & 0.27 & 0.66 & 0.059 & River      & 0.0  & 0.0   & 0.24 & 0.75 & 0.0099 \\
    Vegetation & 0.0  & 0.0    & 0.0  & 0.0  & 1     & Vegetation & 0.0  & 0.0   & 0.0  & 0.02 & 0.98   \\
    \toprule
    \textbf{(E)} & & & \textbf{Confusion Matrix} & & & \textbf{(F)} & & & \textbf{Confusion Matrix} &  \\
     & & &    \textbf{Late Fusion (mean)}    & & & & & &   \textbf{Late Fusion (weighted)} &  \\
    \toprule
    & City & Coastline & Lake & River & Vegetation & & City & Coastline & Lake & River & Vegetation \\
    \midrule
    City       & 1.0   & 0.0    & 0.0  & 0.0   & 0.0    & City       & 0.92 & 0.0  & 0.0   & 0.0   & 0.081  \\
    Coastline  & 0.0   & 0.95   & 0.05 & 0.0   & 0.0    & Coastline  & 0.0  & 0.9  & 0.09  & 0.01  & 0.0    \\
    Lake       & 0.0   & 0.17   & 0.74 & 0.071 & 0.02   & Lake       & 0.0  & 0.15 & 0.71  & 0.12  & 0.02   \\
    River      & 0.0   & 0.0099 & 0.28 & 0.7   & 0.0099 & River      & 0.0  & 0.02 & 0.059 & 0.91  & 0.0099 \\
    Vegetation & 0.059 & 0.0    & 0.0  & 0.089 & 0.85   & Vegetation & 0.03 & 0.0  & 0.0   & 0.099 & 0.87   \\
    \bottomrule
    \end{tabular}
    }
    \caption{Confusion matrix for Sentinel-2 classification (A), Sentinel-1 classification (B), Early Fusion Classification (C), Joint Fusion Classification (D), Late Fusion (mean) Classification (E) and Late Fusion (weighted) Classification (F). }
    \label{tab:confusion}
\end{table*}
Overall, six models (two models, based on the same CNN structure and trained independently on Sentinel-1 and Sentinel-2, and four data fusion models) have been implemented and tested on the validation dataset $\Vec{\mathbf{X}}_{inputs}^v, \Vec{\mathbf{y}}_{ground\_ truths}^v$, through different metrics. For each model, the confusion matrix has been computed and reported in Table \ref{tab:confusion}. By querying this table, it is possible to see how complex classifications for Sentinel-1 (e.g. Vegetation class) or Sentinel-2 (e.g. Lake class) are over-passed by one of the four data fusion paradigms (Early, Joint, Late (mean) and Late (weighted)).

\begin{figure*}[!ht]
    \centering
    \includegraphics[width=1.85\columnwidth]{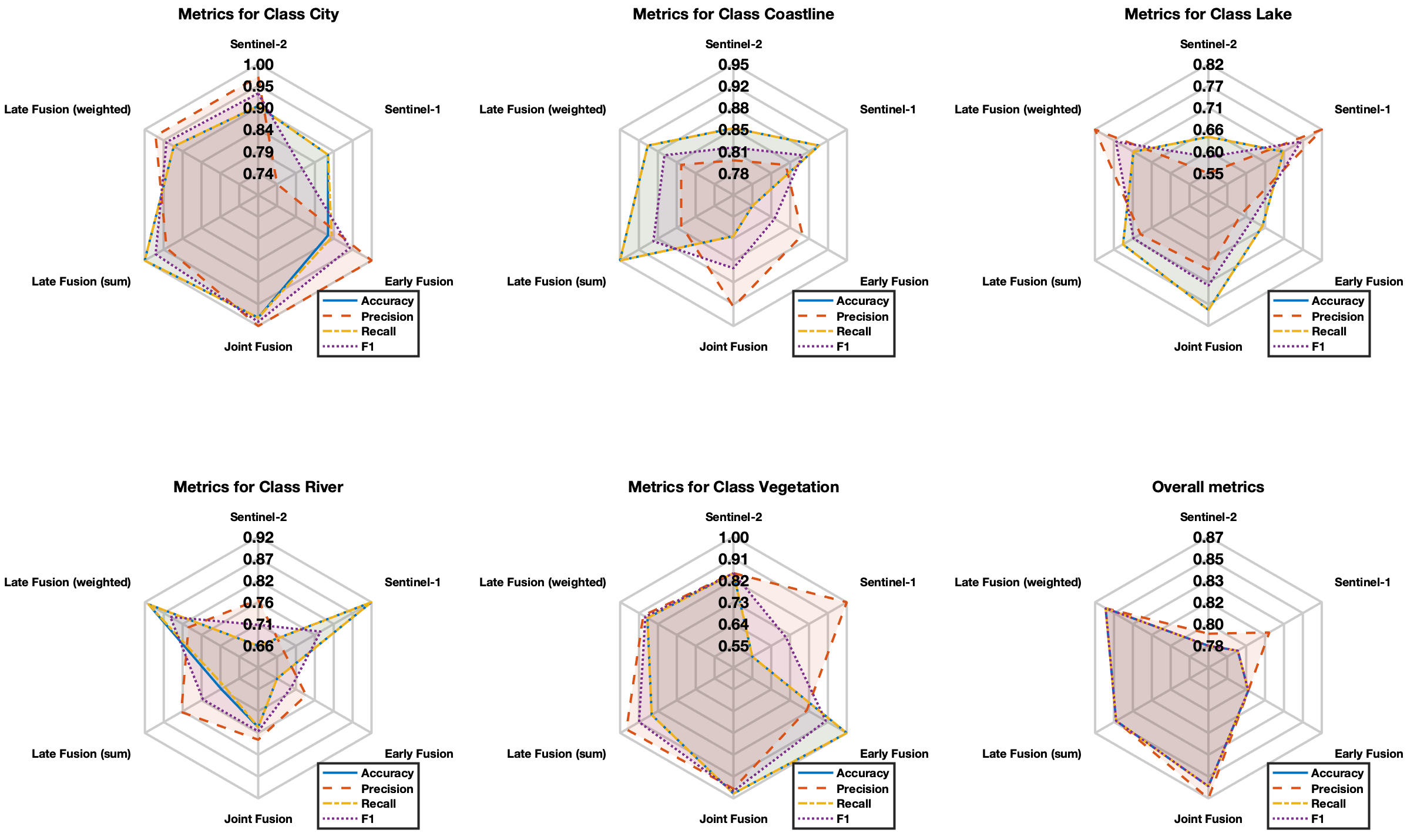}
    \caption{Evaluation metrics: visual representation of Table \ref{tab:metrics}}
    \label{fig:img_metrics}
\end{figure*}

It is also important to note that the data fusion frameworks are not only able to carry out complex classifications, but they are also able to significantly increase the performances of the classification itself. Moreover, this aspect is even more evident if classical metrics used in classification,  such as \textit{accuracy}, \textit{precision}, \textit{recall} and \textit{F1 score}, are used. Accuracy, as a simple ratio of correctly predicted observations over the total number of observations, is a good performance indicator only with symmetric datasets, where values of false positives and false negatives are almost the same. Precision, that is the ratio of correctly predicted positive observations to the total predicted positive observations, shows high values associated to low false positive rates. Recall (or sensitivity) is calculated as the ratio of correctly predicted positive observations over all the observations in the actual considered class, and F1 Score is the weighted average of Precision and Recall, therefore, by taking into account both false positives and false negatives. By using the above metrics as expressed by the following equations:
\begin{equation}
\centering
\resizebox{0.8\columnwidth}{!}{
\begin{tabular}{cc}
   $ A = \frac{Tp+Tn}{Tp+Fp+Fn+Tn}$ & $P = \frac{Tp}{Tp+Fp}$ \\
    & \\
   $R = \frac{Tp}{Tp+Fn}$ & $F1 = 2*\frac{R*P}{R+P}$
\end{tabular}
}
\label{eqn:accuracy}
\end{equation}
for the performance evaluation of the different frameworks,  it is evident (see Table \ref{tab:metrics}) that the performances increase and, particularly for the application selected in this case (land-cover classification), the best data fusion framework is the weighted Late Data Fusion paradigm, both in terms of confusion matrix and metrics. For a better interpretation, data from Table \ref{tab:metrics} have been also reproduced in Fig. \ref{fig:img_metrics}.
\begin{table}[!ht]
    \centering
    \resizebox{0.9\columnwidth}{!}{%
    \begin{tabular}{c|ccccc|c}
    \toprule
	\textbf{Metric} & \textbf{Class 1:} & \textbf{Class 2:} & \textbf{Class 3:} & \textbf{Class 4:} & \textbf{Class 5:} & \textbf{Average} \\
	 & \textbf{City} & \textbf{Coastline} & \textbf{Lake} & \textbf{River} & \textbf{Vegetation} & \\
	\midrule
	& & & \textbf{(A) Sentinel-2} & & &\\
	\midrule
    Accuracy  & 0.90 & 0.85 & 0.64 & 0.66 & 0.85 & 0.78 \\
    Precision & 0.97 & 0.80 & 0.55 & 0.77 & 0.85 & 0.79 \\
    Recall    & 0.90 & 0.85 & 0.64 & 0.66 & 0.85 & 0.78 \\
    F1 Score  & 0.93 & 0.82 & 0.59 & 0.71 & 0.85 & 0.78 \\
	\midrule
	& & & \textbf{(B) Sentinel-1} & & &\\
	\midrule
    Accuracy  & 0.88 & 0.90 & 0.71 & 0.92 & 0.55 & 0.79 \\
    Precision & 0.74 & 0.84 & 0.82 & 0.68 & 1.00 & 0.82 \\
    Recall    & 0.88 & 0.90 & 0.71 & 0.92 & 0.55 & 0.79 \\
    F1 Score  & 0.81 & 0.87 & 0.76 & 0.78 & 0.71 & 0.79 \\
	\midrule
	& & &\textbf{(C) Early Fusion} & & &\\
	\midrule
    Accuracy  & 0.88 & 0.78 & 0.65 & 0.66 & 1.00 & 0.80 \\
    Precision & 1.00 & 0.87 & 0.59 & 0.74 & 0.81 & 0.80 \\
    Recall    & 0.89 & 0.78 & 0.65 & 0.66 & 1.00 & 0.80 \\
    F1 Score  & 0.94 & 0.82 & 0.62 & 0.70 & 0.90 & 0.80 \\
	\midrule
	& & & \textbf{(D) Joint Fusion} & & &\\
	\midrule
    Accuracy  & 0.98 & 0.81 & 0.78 & 0.75 & 0.98 & 0.86 \\
    Precision & 1.00 & 0.92 & 0.68 & 0.78 & 0.96 & 0.87 \\
    Recall    & 0.98 & 0.81 & 0.78 & 0.75 & 0.98 & 0.86 \\
    F1 Score  & 0.99 & 0.86 & 0.72 & 0.76 & 0.97 & 0.86 \\
	\midrule
	& & &    \textbf{(E) Late Fusion} & & &\\
	& & & \textbf{(aggregation: sum)} & & &\\
	\midrule
    Accuracy  & 1.00 & 0.95 & 0.74 & 0.71 & 0.85 & 0.85 \\
    Precision & 0.94 & 0.84 & 0.69 & 0.82 & 0.97 & 0.85 \\
    Recall    & 1.00 & 0.95 & 0.74 & 0.70 & 0.85 & 0.85 \\
    F1 Score  & 0.97 & 0.89 & 0.71 & 0.76 & 0.91 & 0.85 \\
	\midrule
	& & & \textbf{(F) Late Fusion} & & &\\
	& & & \textbf{(aggregation: weighted)} & & &\\
	\midrule
    Accuracy  & 0.92 & 0.90 & 0.71 & 0.91 & 0.87 & 0.86 \\
    Precision & 0.97 & 0.84 & 0.82 & 0.80 & 0.89 & 0.86 \\
    Recall    & 0.92 & 0.90 & 0.71 & 0.91 & 0.87 & 0.86 \\
    F1 Score  & 0.94 & 0.87 & 0.76 & 0.85 & 0.88 & 0.86 \\
    \bottomrule
    \end{tabular}
    }
    \caption{Evaluation metrics for Sentinel-2 classification (A), Sentinel-1 classification (B), Early Fusion Classification (C), Joint Fusion Classification (D), Late Fusion (mean) classification (E) and Late Fusion (weighted) classification (F)}
    \label{tab:metrics}
\end{table}

\vspace{-7pt}
\section{Conclusions}
In this letter, four data fusion paradigms have been investigated (Early, Joint, and two versions of Late Data Fusion), demonstrating that a correct choice of the data fusion framework can bring better results. Indeed, by fixing a basic structure for the classification model, as that represented in Figure 4, and by varying the data fusion paradigm, the proposed procedure  demonstrates that, in the case of land-cover classification, the Late Data Fusion (with a weighted strategy) leverages to better results than the others. 
Therefore, the main strengths of this letter are to have explored and demonstrated that the Late Data Fusion (with a weighted strategy) is the best choice for land-cover classification, and help interested researchers in their work when data fusion applied to remote sensing is involved.  

Since different versions of the three proposed Data Fusion paradigms can be designed, among the future works, there is a further exploration of the proposed procedure when other Data Fusion paradigms will be examined. Moreover, this methodology can be also validated for other  AI4EO applications (e.g. segmentation or change detection), to measure its impact on final results.
\vspace{-10pt}
\small
\section*{Acknowledgments}
This research is supported by the Open Space Innovation Platform (OSIP) project titled "Al powered cross-modal adaptation techniques applied to Sentinel-1 and -2 data" under a joint collaboration between the European Space Agency (ESA) $\Phi$-Lab and the University of Sannio.
\vspace{-10pt}
\balance
\bibliographystyle{IEEEtran}
\bibliography{IEEEabrv,refs}
\end{document}